\documentclass[11pt]{article}
\usepackage{times,latexsym}
\usepackage{url}
\usepackage[T1]{fontenc}
\usepackage[preprint]{acl}
\usepackage{booktabs}
\usepackage{tabularx,array}
\usepackage{float}
\usepackage{graphicx}
\usepackage{amsmath,amssymb}
\usepackage{tikz}
\usetikzlibrary{positioning,arrows.meta,fit,backgrounds}

\newcommand{\msr}{Exact}

\newtheorem{definition}{Definition}

\title{Procedure without Judgment:\\
What Small Language Models Learn under an Exact Verifier}
\author{Arunkumar V \\ Anna University \\ Tiruchirappalli, India \\ \texttt{arunkumarv1530@gmail.com} \And Manoranjan Gandhudi \\ Central University of Karnataka \\ Kalaburagi, India \And Gangadharan G. R. \\ NIT Tiruchirappalli \\ Tiruchirappalli, India \AND Arun Prakash \\ Jawaharlal Nehru University \\ New Delhi, India \And S. Senthilkumar \\ Anna University \\ Tiruchirappalli, India}

\begin{document}
\maketitle

\begin{abstract}
When an authorization policy blocks an agent's tool call, the audit log should name the facts
that forced the decision and nothing more. For an executable policy over an enumerable context space
this minimal sufficient reason can be computed exactly, so an explanation is scored against ground
truth rather than a proxy for it. We use that exactness to ask what a model can learn from a verifier
that has no error in it. The explainer sees one request and one decision, never the policy, and has
thirty turns in which it may call the policy on altered requests and ask whether a candidate set of
facts is sufficient. Frontier models find the exact reason about half the time with the tools alone,
and one sentence telling them to shrink the answer lifts Haiku 4.5 from 58\% to 75\%. A 4B model
fine-tuned on 2000 traces of the deletion algorithm reaches 92\% on the policy it was trained on.
Teacher kind is what moves this: at matched data size the algorithm-taught student leads the
frontier-taught one in all three training seeds, and the two arms never overlap. What the student
learns is a fixed schedule of sufficiency checks, and the schedule does not travel. On three public
policies it never saw, its answers stay sufficient (94\%) and stop being minimal (48\%), while Haiku
under the same sentence is exact on all 47 episodes. Reinforcement learning from the exact reward
leaves held-out accuracy where it was and inverts the strategy, from probing the policy to spending
the entire budget on sufficiency checks. The oracle made the effect of training legible, but it did
not make training better.
\end{abstract}

\section{Introduction}
\label{sec:intro}
Reinforcement learning from verifiable rewards works where a verifier exists, and the field is now
asking what happens when the verifier is a rubric, a judge, or a heuristic that only approximates the
objective \citep{gamingverifiers2026,rubrichacking2026}. We take the other end of that question and
suppose the verifier is exact. What can a model learn from it, and what can it not?

Explaining a decision of an executable authorization policy is a task where exactness is available
and the answer is non-trivial. An agent asks to send a file to an external
domain, and the policy blocks it. A useful audit entry names the facts that forced the block, say the
target domain and the absence of elevated authorization, and leaves out the facts that did not
matter. In the formal explanation literature this is an abductive explanation or prime implicant
\citep{shih2018symbolic,ignatiev2019abduction}, a subset of the request's features whose values
entail the decision under every completion of the rest, with no proper subset that does. We call it
a minimal sufficient reason (MSR). For a policy over an enumerable context space an oracle can enumerate every MSR of every
decision, so a cited reason set is scored against ground truth rather than a proxy for it.

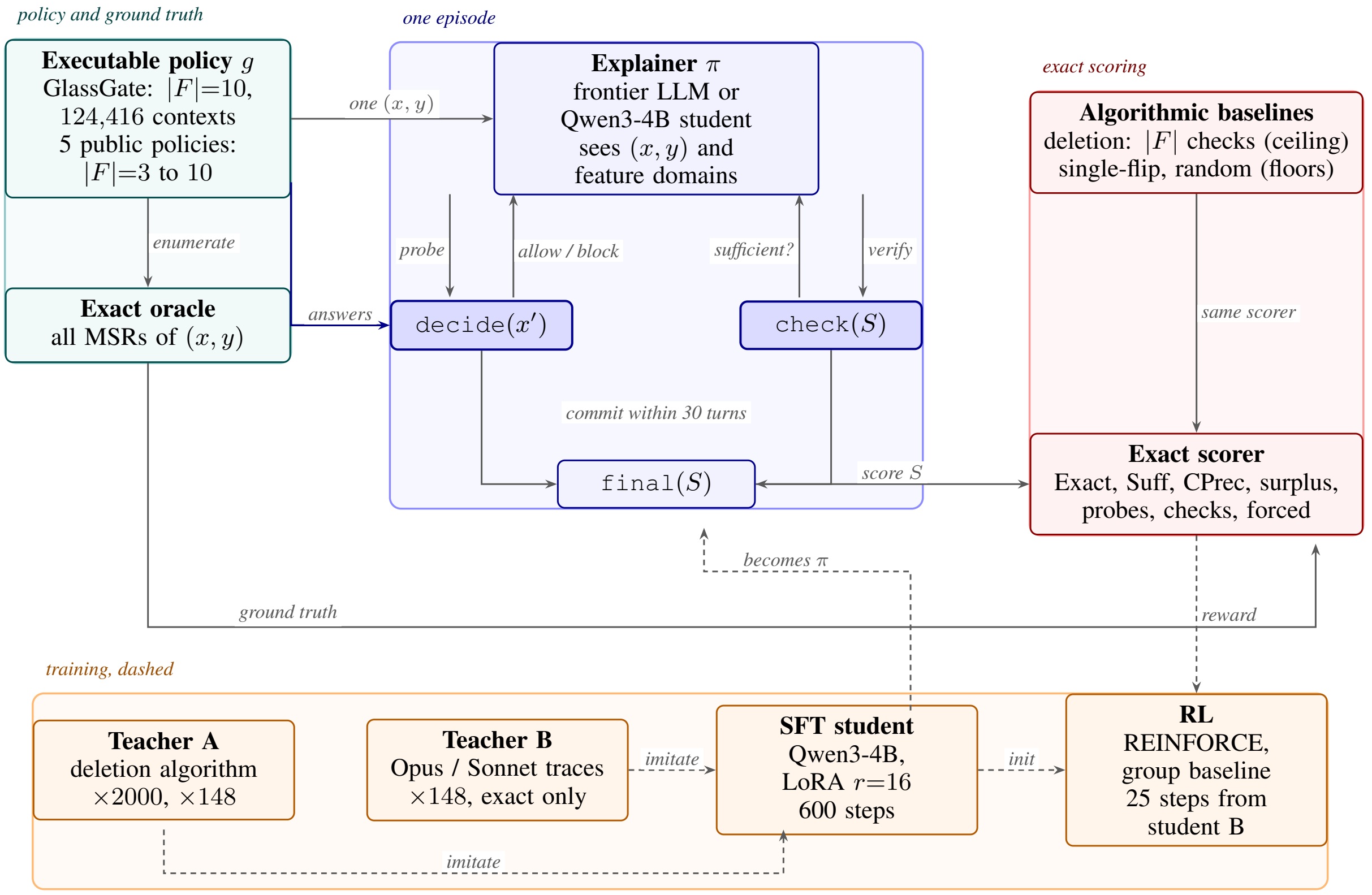
\begin{figure*}[t]
  \centering
  \resizebox{\textwidth}{!}{
\begin{tikzpicture}[
  font=\footnotesize, x=1mm, y=1mm,
  >={Stealth[length=4.2pt]},
  panel/.style={rounded corners=4pt, line width=0.7pt, inner sep=0pt},
  ptitle/.style={font=\scriptsize\itshape, inner sep=2pt, anchor=west},
  gbox/.style={draw=teal!60!black,   fill=teal!5,    rounded corners=2.5pt, align=center, inner sep=4pt,
            execute at begin node={\hyphenpenalty=10000\exhyphenpenalty=10000\relax}, line width=0.6pt},
  bbox/.style={draw=blue!55!black,   fill=blue!5,    rounded corners=2.5pt, align=center, inner sep=4pt,
            execute at begin node={\hyphenpenalty=10000\exhyphenpenalty=10000\relax}, line width=0.6pt},
  tbox/.style={draw=blue!55!black,   fill=blue!14,   rounded corners=2.5pt, align=center, inner sep=4pt,
            execute at begin node={\hyphenpenalty=10000\exhyphenpenalty=10000\relax}, line width=0.8pt},
  rbox/.style={draw=red!55!black,    fill=red!5,     rounded corners=2.5pt, align=center, inner sep=4pt,
            execute at begin node={\hyphenpenalty=10000\exhyphenpenalty=10000\relax}, line width=0.6pt},
  ybox/.style={draw=orange!70!black, fill=orange!8,  rounded corners=2.5pt, align=center, inner sep=4pt,
            execute at begin node={\hyphenpenalty=10000\exhyphenpenalty=10000\relax}, line width=0.6pt},
  lbl/.style={font=\scriptsize\itshape, text=black!65, inner sep=1.5pt, fill=white},
  arr/.style={->, line width=0.6pt, draw=black!65},
  darr/.style={arr, dashed, dash pattern=on 2pt off 1.6pt},
  barr/.style={arr, draw=blue!55!black, line width=0.7pt},
]

\node[gbox, text width=33mm] (policy) at (20,0)
  {\textbf{Executable policy} $g$\\ GlassGate: $|F|{=}10$,\\ $124{,}416$ contexts\\ 5 public policies: $|F|{=}3$ to $10$};
\node[gbox, text width=33mm] (oracle) at (20,-26)
  {\textbf{Exact oracle}\\ all MSRs of $(x,y)$};
\draw[arr] (policy) -- node[lbl,right] {enumerate} (oracle);

\node[bbox, text width=38mm] (llm) at (84,0)
  {\textbf{Explainer} $\pi$\\ frontier LLM or Qwen3-4B student\\ sees $(x,y)$ and feature domains};
\node[tbox, text width=20mm] (decide) at (62,-26) {\texttt{decide}$(x')$};
\node[tbox, text width=20mm] (check)  at (106,-26) {\texttt{check}$(S)$};
\node[bbox, text width=22mm] (final)  at (84,-46) {\texttt{final}$(S)$};

\draw[arr] (58,-9.5) -- node[lbl,left,pos=.55] {probe} (58,-22.5);
\draw[arr] (66,-22.5) -- node[lbl,right,pos=.45] {allow / block} (66,-9.5);
\draw[arr] (110,-9.5) -- node[lbl,right,pos=.55] {verify} (110,-22.5);
\draw[arr] (102,-22.5) -- node[lbl,left,pos=.45] {sufficient?} (102,-9.5);

\draw[arr] (decide.south) |- (final.west);
\draw[arr] (check.south)  |- (final.east);
\node[lbl] at (84,-37) {commit within 30 turns};

\node[rbox, text width=39mm] (base) at (152,-3)
  {\textbf{Algorithmic baselines}\\ deletion: $|F|$ checks (ceiling)\\ single-flip, random (floors)};
\node[rbox, text width=39mm] (scorer) at (152,-46)
  {\textbf{Exact scorer}\\ \msr, Suff, CPrec, surplus,\\ probes, checks, forced};
\draw[arr] (base) -- node[lbl,right] {same scorer} (scorer);
\draw[arr] (final.east) -- node[lbl,above] {score $S$} (scorer.west);

\draw[arr]  (policy.east) -- node[lbl,above] {one $(x,y)$} (llm.west);
\draw[barr] ([yshift=-8mm]policy.east) |- node[lbl,above,pos=.75] {answers} (decide.west);
\draw[arr] (oracle.south) -- (20,-64) -- node[lbl,above,pos=.12] {ground truth} (167,-64) -- (167,-53.5);

\node[ybox, text width=30mm] (tA)  at (22,-82)
  {\textbf{Teacher A}\\ deletion algorithm\\ $\times 2000$, $\times 148$};
\node[ybox, text width=30mm] (tB)  at (64,-82)
  {\textbf{Teacher B}\\ Opus / Sonnet traces\\ $\times 148$, exact only};
\node[ybox, text width=30mm] (sft) at (108,-82)
  {\textbf{SFT student}\\ Qwen3-4B, LoRA $r{=}16$\\ 600 steps};
\node[ybox, text width=30mm] (rl)  at (152,-82)
  {\textbf{RL}\\ REINFORCE, group baseline\\ 25 steps from student B};

\draw[darr] (tB.east) -- node[lbl,above] {imitate} (sft.west);
\draw[darr] (sft.east) -- node[lbl,above] {init} (rl.west);
\coordinate (pad) at (60,-97);
\draw[darr] (22,-89.5) -- (22,-95) -- node[lbl,above] {imitate} (100,-95) -- (100,-89.5);
\draw[darr] (116,-74.5) -- (116,-57) -- node[lbl,above,pos=.6] {becomes $\pi$} (90,-57) -- (90,-51.5);
\draw[darr] (scorer.south) -- node[lbl,right] {reward} (rl.north);

\begin{scope}[on background layer]
  \node[panel, draw=teal!45,   fill=teal!3,   fit=(policy)(oracle)] (P1) {};
  \node[panel, draw=blue!45,   fill=blue!3,   fit=(llm)(decide)(check)(final)] (P2) {};
  \node[panel, draw=red!40,    fill=red!3,    fit=(base)(scorer)] (P3) {};
  \node[panel, draw=orange!55, fill=orange!4, fit=(tA)(tB)(sft)(rl)(pad)] (P4) {};
\end{scope}
\node[ptitle, text=teal!50!black]   at ([shift={(1mm,2.8mm)}]P1.north west) {policy and ground truth};
\node[ptitle, text=blue!45!black]   at ([shift={(1mm,2.8mm)}]P2.north west) {one episode};
\node[ptitle, text=red!45!black]    at ([shift={(1mm,2.8mm)}]P3.north west) {exact scoring};
\node[ptitle, text=orange!60!black] at ([shift={(1mm,2.8mm)}]P4.north west) {training, dashed};
\end{tikzpicture}}
  \caption{\textbf{Explanation as experiment.} An explainer sees a context $x$ and the policy's decision $y$
  but never the policy. It may call \texttt{decide} on any context and \texttt{check} on any candidate
  reason set, at most $k{=}30$ times, then commits to a reason set $S$. An exact oracle over the enumerable
  context space knows every minimal sufficient reason, so $S$ is scored without a proxy. The deletion
  algorithm, which needs exactly $|F|$ \texttt{check} calls, is the ceiling. The same episode format and the
  same scorer serve frontier models, the supervised students and RL, so every number in the paper is on
  a shared set of episodes. Bottom: the procedure is installed three ways, from algorithmic
  traces, from frontier traces, and by RL from the scorer's reward. Blue marks the two tools and the arrow carrying the policy's answers.}
  \label{fig:arch}
\end{figure*}

The model does not see the policy. It sees one request, one decision, and two tools. It may call the
policy on any altered request, and it may ask whether a candidate set of features is sufficient. It
has thirty turns. Finding the MSR under those rules is an experiment, and this paper asks who can run
it. Figure~\ref{fig:arch} shows the setup.

Four results follow. First, frontier accuracy does not follow model tier, and one sentence of
instruction moves it. On 120 shared
episodes with both tools, Haiku 4.5 reaches the exact MSR on 58\% and Sonnet 5 on 50\%, against 16\% for Gemini Flash-Lite. Opus 5 is exact on all 21 episodes it did not refuse, but that
subset is easier and we do not read it as a ceiling (Section~\ref{sec:frontier}). Adding a single
instruction to remove any cited feature whose removal keeps the decision lifts Haiku to 75\% and
Flash-Lite to 43\%. Sonnet does not move, for a reason unrelated to what it cites. On 55 of 120 episodes it
answers in prose instead of the required action format and the harness ends the episode empty; where
it commits it is exact 94\% of the time. Its score is a
format-compliance number, and we report it as one (Section~\ref{sec:frontier}).

Second, the deletion algorithm is a better teacher for a small model than a frontier model is. We
fine-tune a small language model, Qwen3-4B, on traces of the algorithm and, separately, on traces of
Opus and Sonnet solving the same episodes, keeping only their exact answers so that both teachers are
correct on every training episode. At matched data size the algorithm-taught student is ahead in every one of three training seeds, by
16, 47 and 61 points, with the worst algorithm seed above the best frontier seed. With 2000 traces it reaches 92\%, 39 points ahead of the
frontier-taught student ($p<10^{-4}$). It is also 11 points ahead of Haiku under the same shrink instruction
(81\%), a gap these 64 episodes cannot resolve ($p=0.12$), and uses fewer calls: ten \texttt{check}s and no probes against Haiku's 8.6 checks and 8.8 probes (Section~\ref{sec:teaching}).

Third, what the student learned is a schedule, and the schedule does not travel. On three public
policies written in OPA and Cedar, with three to ten features, the students keep producing sufficient
reasons (94 to 98\%) and stop producing minimal ones (down 29 to 44 points). Haiku with the shrink
instruction, evaluated on the same 47 episodes, is exact on all of them. The student's advantage is
confined to the gate it was trained on, so its in-distribution lead is not evidence about small-model
capability in general (Section~\ref{sec:transfer}).

Fourth, reinforcement learning from the exact reward moves the strategy and not the accuracy. Warm-started from the
frontier-taught student and trained with a group-relative baseline against a reward in which exact
minimality dominates, the student does not get more accurate on held-out episodes ($+3$ points, $p=0.80$).
It changes how it works, stopping probing entirely and spending its whole budget on sufficiency
checks, which is the deletion algorithm's query shape without its accuracy. A training-distribution
gain vanished on the held-out set, and a second warm start diverged. Padding under a sufficiency-heavy
reward and starvation at the reward ceiling both appear with the verifier held exact, which places them in the reward's shape rather than in its error (Section~\ref{sec:rl}).

Every cross-system number is computed on the intersection of
episodes the compared systems share, under the same prompt, with $n$ stated.

\section{Setting: explaining an executable policy by querying it}
\label{sec:setting}
\paragraph{Policies and contexts.}
GlassGate is an executable authorization policy for an agent gateway \citep{camel2025,agentdojo2024}. A request is a context $x$ over
ten categorical features: the tool being called (twelve values), the scope level, the target domain,
three content flags for personal, credential and financial data, the caller's authorization level and
role, the declared intent, and a count of prior violations. The product of the domains is
$124{,}416$ contexts, and the policy is a fixed function $g: X \to \{\text{allow}, \text{block}\}$.
Because $X$ is enumerable, every claim below about sufficiency is decided by evaluating $g$, not by
sampling. Five public policies are the held-out domains: a load-balancer policy in OPA, two Kubernetes
Gatekeeper constraints (seccomp profiles and container limits), and two Cedar policies (a streaming
service and a tax-preparation service). They have three to ten features and run on the same oracle
machinery. The students were scored on the OPA and Cedar policies, the frontier models on all five. All episodes are stratified over the decision and over MSR size.

\paragraph{What the policy is.}
GlassGate is nine rules checked in order with the first match deciding, eight BLOCK rules over a final
ALLOW default. Each BLOCK rule is a conjunction of two or three literals over the ten features.
\texttt{R1 credential-egress}, for instance, blocks when the payload contains credentials, the target
domain is \texttt{public.web} or \texttt{unknown.xyz}, and the tool is an outbound channel
(\texttt{email.send}, \texttt{http.post} or \texttt{files.write}); Appendix~\ref{app:policy} lists all
nine. The rules were written to overlap, so a context usually carries distractors: literals that
pattern-match some BLOCK rule's antecedent, and so look risk-relevant, while lying outside every
minimal sufficient reason. Blocked contexts carry $3.55$ such distractors on
average, $91\%$ carry at least two, and $52\%$ of contexts admit more than one minimal sufficient
reason. Blocks are $54.6\%$ of the space. The policy was fixed before any model reported here was run
and was not changed afterwards. The two
decisions are asymmetric. A block is sufficient as soon as one rule's antecedent is pinned, so its
minimal reasons run two to three features, while an allow has to defeat all eight BLOCK rules at once
and runs three to six (Appendix~\ref{app:failure}).

\paragraph{Minimal sufficient reasons.}\nopagebreak
\begin{definition}
For a context $x$ with decision $y = g(x)$, a feature set $S \subseteq F$ is \emph{sufficient} if every
context $x'$ that agrees with $x$ on $S$ satisfies $g(x') = y$. It is a \emph{minimal sufficient reason}
if it is sufficient and no proper subset is.
\end{definition}
This is the abductive explanation of \citet{ignatiev2019abduction} and the prime implicant of
\citet{shih2018symbolic}, specialised to a Boolean decision. A decision may have several MSRs of
different sizes. The oracle enumerates all of them.

\paragraph{The episode.}
The explainer receives $x$, $y$, the feature names and their domains, and a preamble describing an
audit-log task. It never receives $g$. Each turn it may issue one of three actions.
\texttt{decide}$(x')$ returns $g(x')$ for any context $x'$. \texttt{check}$(S)$ returns whether $S$ is
sufficient for $(x,y)$. \texttt{final}$(S)$ commits to $S$ and ends the episode. The budget is $k=30$
turns. A model that has not committed by then is asked once for its answer, and we record that it was
forced. The \texttt{check} tool is what makes an exact minimality search possible in $|F|$ queries,
and we give it to every system in the paper, so no comparison turns on who had which tool.

\paragraph{Metrics.}
Given the committed set $S$ and the oracle's MSR set $\mathcal{M}$, we report the fraction of episodes
on which $S \in \mathcal{M}$ (\msr, the headline), the fraction on which $S$ is sufficient (Suff),
the precision of $S$ against the union of all MSRs (CPrec), the surplus $|S| - \min_{M \in
\mathcal{M}} |M|$ (positive means padding, negative means an insufficient answer), and the mean number
of \texttt{decide} and \texttt{check} calls per episode, which describe the strategy. Every
comparison between two systems is a paired sign-flip permutation test on the per-episode \msr\
difference over the episodes both systems ran, with $20{,}000$ permutations.

\paragraph{Ceiling and floors.}
The deletion algorithm starts from $S=F$ and removes each feature that is not needed for sufficiency,
returning an MSR in exactly $|F|$ \texttt{check} calls with no probes, so it is the ceiling for any
explainer with the same tools. Its surplus in Table~\ref{tab:main} is positive with \msr\ at 1.0 because
it returns an MSR, not necessarily the smallest. Single-flip cites the features whose individual change
flips the decision; random probing cites what it saw co-vary over random contexts. These bound what tool
access alone achieves.

\paragraph{Systems.}
Frontier explainers are Gemini 3.5 Flash-Lite, Claude Haiku 4.5, Claude Sonnet 5 and Claude Opus 5,
thinking disabled, both tools. Opus declined most episodes as a security task, the refusal bias documented by
\citet{defensiverefusal2026}, so its numbers are on the 21 of 120 it answered. Students are
Qwen3-4B \citep{qwen3} with a rank-16 LoRA \citep{hu2022lora}, 600 steps, one recipe for all arms. Each student is named for its
teacher and its number of training traces. Alg-2000 and Alg-148 are taught by the deletion algorithm.
Frontier-148 is taught by Opus 5 or Sonnet 5 traces, kept only where the model committed to an exact
answer. Alg-public is taught by the deletion algorithm on the five public policies only, never on
GlassGate (Section~\ref{sec:transfer}). The RL arm warm-starts from Frontier-148 and runs group-relative
REINFORCE \citep{williams1992reinforce,shao2024grpo} for 25 steps. Appendix~\ref{app:hparams} has every setting.

\section{Can frontier models find the minimal reason?}
\label{sec:frontier}
\begin{figure*}[t]
  \centering
  \includegraphics[width=\textwidth]{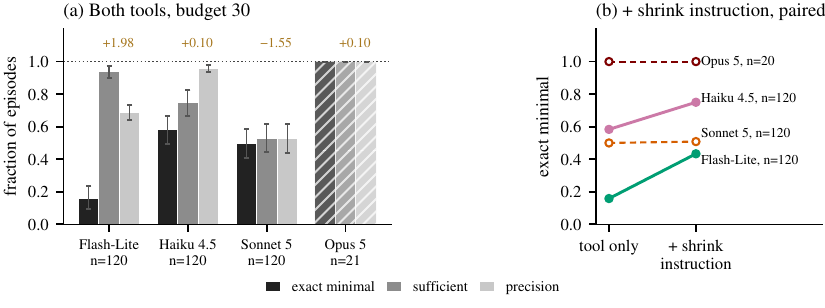}
  \caption{\textbf{Frontier models with both tools, and what one sentence changes.} (a) Exact
  minimality, sufficiency and precision on GlassGate with both tools and no further instruction, all
  three models on the same 120 episodes and Opus on the 21 it did not refuse (hatched), a subset with no minimal reason above three features. The number above
  each group is the mean surplus of cited features over the smallest MSR; Sonnet's is negative because
  46\% of its episodes end in a forced empty answer (Section~\ref{sec:frontier}). The dotted line is the
  deletion algorithm. (b) Paired within model on the episodes each model ran under both prompts ($n$ on the plot; Opus's shrink arm has 20), with and without a one-sentence shrink instruction. Solid lines are gains of more than two points.}
  \label{fig:tiers}
\end{figure*}

We first ask what models never trained on the task do with the two tools. Figure~\ref{fig:tiers} shows
four frontier models on GlassGate at budget 30, without and with the shrink instruction;
Appendix~\ref{app:frontier} has every domain and Appendix~\ref{app:failure} the breakdown by reason size.

\paragraph{Two tiers, then one sentence.}
With the tools alone, Haiku 4.5 reaches the exact MSR on 58\% of episodes and Flash-Lite on 16\%, a gap
of 42 points ($p<10^{-4}$, paired). Sonnet 5, which one would expect above Haiku, is at 50\%, eight
points below it ($p=0.064$). Opus 5 is exact on all 21 episodes it did not refuse, 24 points above Sonnet there ($p=0.063$, $n=21$),
but that subset is easier: its minimal reasons average 2.29 features against 3.02, it contains none of the
32 episodes whose reason exceeds three, and Haiku and Sonnet both reach 76\% on it. Opus is not
established as a ceiling. One added sentence, telling the model to remove any cited feature
whose removal keeps the decision, lifts Flash-Lite 28 points to 43\% and Haiku 17 to 75\% (both
$p\le10^{-4}$, paired within model, Figure~\ref{fig:tiers}b). It repairs the failure both share, citing
one feature too many: Haiku's sufficiency goes from 75\% to 83\% and its surplus falls.

\paragraph{Sonnet's score is a format number.}
Sonnet does not move under the instruction (50\% to 51\%), and the reason is visible in the trajectories.
On 55 of 120 episodes in each arm (45 in both), Sonnet twice replies with reasoning prose instead of the required single JSON action, and the harness ends the episode with an empty answer scoring zero on every metric. It happens at turn 20 on average, not at the budget, and on the hard episodes: 17 of the 18 whose smallest reason has four features are among the 55. On the 65 where it commits it is exact 92\% without the instruction and 94\% with it, against Haiku's 95\% on those same episodes, while its overall sufficiency, 53\%, sits below Flash-Lite's 94\%. Haiku, Flash-Lite and Opus have at most one forced episode each. We report the forced rate rather than reinterpret the score; a more forgiving harness would change Sonnet's numbers and no one else's.

\section{Teaching a small language model the procedure}
\label{sec:teaching}
Can a small language model \citep{belcak2025slm} be taught what the frontier does with the instruction? We fine-tune one
student, Qwen3-4B, from three teachers with one recipe, and evaluate all of them on the same 64 GlassGate
episodes with the same tools, the same budget and the same shrink instruction, against Haiku and Sonnet
under that same instruction. Table~\ref{tab:main} gives the numbers and Table~\ref{tab:pairs} the
paired tests.

\paragraph{Teachers.}
The algorithm teacher is the deletion procedure of Section~\ref{sec:setting}, rendered as an episode
with a \texttt{check} call per feature, a one-line rationale for each, and a \texttt{final}. It
supplies 2000 training episodes for Alg-2000 and 148 for Alg-148. The frontier teacher is Opus 5 or
Sonnet 5 solving 148 episodes with the same tools and the same prompt, and we keep only the
trajectories that committed on their own to an exact minimal reason, so Frontier-148 sees nothing but
correct answers. Both teachers are correct on every training episode. They differ in kind, and
Alg-148 against Frontier-148 is the pair that holds volume fixed and isolates kind. No teacher and no student ever sees the policy. The untrained base model with the shrink
prompt is the first row of Table~\ref{tab:main}, and is what prompting alone achieves at this scale.

\paragraph{The algorithm is the better teacher.}
Alg-2000 reaches the exact MSR on 92\% of episodes, against 53\% for Frontier-148, a difference of 39
points ($p<10^{-4}$), and eight points short of the deletion ceiling ($p=0.064$). Against the frontier
under the same instruction it is 11 points ahead of Haiku, a gap 64 episodes cannot resolve
($p=0.12$), and 36 ahead of Sonnet ($p<10^{-4}$, with Sonnet bounded by its format failures). It gets
there with ten \texttt{check} calls and no probes, against Haiku's 8.6 checks and 8.8 probes.
Frontier-148 does not get there at all: 28 points behind Haiku, $p<10^{-3}$. Volume and teacher kind can be separated, because Alg-148 and Frontier-148 saw the same number of
traces. Alg-148 wins, and it wins in every one of three training seeds
(Appendix~\ref{app:seeds}). Seed variance is wide, 69\% to 97\% against 36\% to 53\%, yet the arms
never overlap: the worst algorithm seed beats the best frontier seed, and that is the claim. On all
120 GlassGate episodes the margin is 17.5 points ($p=0.006$); on the stratified 64 it is 16 points and
does not reach significance ($p=0.11$), so that comparison was underpowered rather than negative.
Going from 148 traces to 2000 adds a further 23 points ($p=0.003$), so kind and volume both contribute
and are separable.

\paragraph{Not everything clears the heuristic floor.}
Single-flip changes one feature at a time and cites what flips the decision, reaching 52\% on
GlassGate with 28 probes and no \texttt{check} calls. Table~\ref{tab:pairs} tests every system against that floor. Alg-2000 clears it by 41 points and
instructed Haiku by 30, both at $p<10^{-3}$. Frontier-148 does not clear it, before or after RL, and
neither does Sonnet with the instruction. Section~\ref{sec:rl}'s RL null is therefore measured on a
policy that starts at the heuristic floor.

\paragraph{The teacher installs a schedule.}
The probes and checks columns of Table~\ref{tab:main} show how each student spends its budget. The
algorithm-taught students never call \texttt{decide}: they run ten \texttt{check} calls, one per
feature, and commit, reproducing the deletion algorithm's query pattern exactly. Frontier-148
probes 3.5 times per episode and checks 25.5 times, its teachers' probe-then-verify order at three
times their check count (Opus and Sonnet averaged seven checks per episode). Each student learned how
to spend thirty turns, with the count anchored to the training gate and not to the policy at hand:
Appendix~\ref{app:failure} shows the algorithm-taught students
making ten checks on the three-feature policy and about fourteen on the seven- and ten-feature ones.
Appendix~\ref{app:example} gives a worked episode.

\begin{table*}[t]
  \centering
  \caption{\textbf{Students, RL and algorithmic baselines on the same 64 GlassGate episodes.} Students are
  named for their teacher and their number of training traces. Suff is
  the fraction of sufficient answers, CPrec the precision against the union of all MSRs, surplus the
  mean number of cited features beyond the smallest MSR (negative means insufficient), and probes and
  checks the mean \texttt{decide} and \texttt{check} calls per episode, forced the fraction of
  episodes ended by the harness after two malformed replies. Every system, including the two frontier
  rows, ran with the shrink instruction. The deletion algorithm's surplus is positive with \msr\ at 1.0
  because it returns an MSR, not necessarily the smallest one. Bold marks the best of the five student rows on \msr, Suff and CPrec.}
  \label{tab:main}
  \small
  \begin{tabular}{lccccccc}
\toprule
system & Exact & Suff & CPrec & surplus & probes & checks & forced \\
\midrule
base 4B (prompt only) & 0.000 & 0.250 & 0.430 & +2.42 & 0.0 & 21.8 & 0.00 \\
Frontier-148 & 0.531 & 0.781 & 0.841 & +1.17 & 3.5 & 25.5 & 0.00 \\
\quad + RL, exact reward & 0.562 & 0.734 & 0.877 & +0.75 & 0.0 & 30.0 & 0.00 \\
Alg-148 & 0.688 & \textbf{1.000} & 0.914 & +0.88 & 0.0 & 10.2 & 0.00 \\
Alg-2000 & \textbf{0.922} & 0.984 & \textbf{0.973} & +0.38 & 0.0 & 10.0 & 0.00 \\
\midrule
Haiku 4.5 + shrink & 0.812 & 0.906 & 0.980 & +0.25 & 8.8 & 8.6 & 0.00 \\
Sonnet 5 + shrink & 0.562 & 0.578 & 0.574 & $-$1.55 & 3.3 & 8.9 & 0.42 \\
\midrule
deletion algorithm (ceiling) & 1.000 & 1.000 & 1.000 & +0.27 & 0.0 & 10.0 & 0.00 \\
single-flip heuristic & 0.516 & 0.516 & 0.750 & $-$1.08 & 28.0 & 0.0 & 0.00 \\
random probing & 0.000 & 0.797 & 0.524 & +3.56 & 30.0 & 0.0 & 0.00 \\
\bottomrule
\end{tabular}

\end{table*}

\section{What transfers to policies the student never saw}
\label{sec:transfer}
\begin{figure*}[t]
  \centering
  \includegraphics[width=\textwidth]{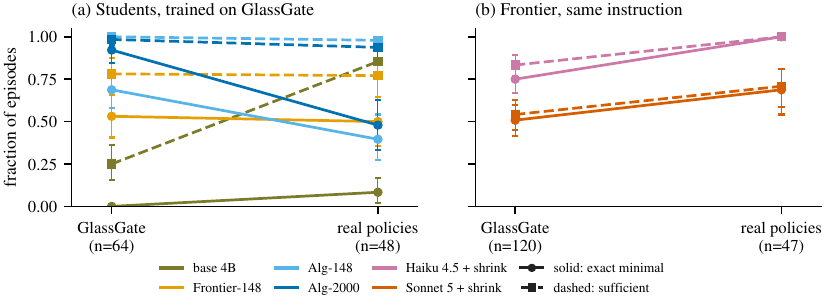}
  \caption{\textbf{What survives the move to policies the student never saw.} Solid lines are exact
  minimality, dashed lines sufficiency, on GlassGate and on the three public policies the students ran.
  (a) The algorithm-taught students keep sufficiency and lose minimality. (b) Haiku and Sonnet with the
  same shrink instruction move both together, upward, because most public policies have fewer features
  than GlassGate. Per-policy numbers with shared-episode counts are in Appendix~\ref{app:offdist}.}
  \label{fig:transfer}
\end{figure*}

The three public policies test whether the students of Section~\ref{sec:teaching} learned the
procedure or the gate. Figure~\ref{fig:transfer} plots exact minimality and sufficiency on GlassGate and on the real policies; Appendix~\ref{app:offdist} has the per-policy numbers.

\paragraph{Sufficiency transfers and minimality does not.}
Alg-2000 keeps its answers sufficient off-distribution, 94\% against 98\% at home, and
Alg-148 likewise, 98\% against 100\%. Their exact minimality falls from 92\% to 48\% and from 69\% to
40\%. The gap between sufficiency and minimality opens by 46 and 58 points on the real policies, both
at $p<10^{-3}$. The verification step, which keeps checking until the set is sufficient, survives the move
to policies with different features and different sizes. The search, which removes features until nothing more can
go, does not. Frontier-148 is flat on both measures pooled, and off-distribution the ordering reverses. In all
three seeds the frontier-taught student is the more minimal one on the public policies, 50\%, 58\%
and 54\% against 40\%, 56\% and 44\%, while the algorithm-taught student keeps the higher
sufficiency, 96\% to 98\% against 77\% to 83\%. That advantage is not general. On
cedar-streaming-service, the only ten-feature public policy, the frontier-taught student is exact on
zero episodes in all three seeds while the algorithm-taught students reach 6\% to 31\%. What travels is citing few features, which pays when the answer has few features.

\paragraph{The frontier models do not show the split.}
Haiku and Sonnet with the same instruction move both quantities together and upward on the same three
policies. Haiku goes from 75\% to 100\% exact and 83\% to 100\% sufficient on all 47 episodes; Sonnet from 51\% to 69\% exact and 54\% to 71\% sufficient, its forced rate falling with the shorter policies. Most public
policies have fewer features than GlassGate, so a general search finds the minimal set more easily and
the untrained models behave as general searchers. The students behave as models that learned a
schedule for ten features, so the split is a signature of the taught procedure and not of the task.

\paragraph{The schedule is policy-specific in both directions.}
Training on GlassGate and testing elsewhere cannot separate learning the procedure from memorising
GlassGate's ten-feature structure, so we ran the mirror. Alg-public is taught by the same deletion algorithm on
the five public policies only, 80 trajectories, never GlassGate. It reaches 92\% on those policies
against 40\% to 50\% for every GlassGate-trained student, so the recipe is not a GlassGate artefact. On
GlassGate it reaches 28\%, above the untrained base (0\%, $p=10^{-4}$) but below the frontier-taught
student ($-25$ points, $p=0.010$), the matched algorithm-taught student ($-41$ points, $p=10^{-4}$) and even
single-flip ($-23$ points, $p=0.021$). Something transfers and it is not enough to be useful: the algorithm
teaches a schedule fitted to the policy it was taught on, whichever direction the experiment runs.

\paragraph{Off-distribution the student is behind.}
On the policies it never saw, Alg-2000 is behind instructed Haiku on all three, by 50, 38 and 73
points, every one significant; Appendix~\ref{app:offdist} has the per-policy numbers. It
is behind Sonnet on two of the three.
The level comparison of Section~\ref{sec:teaching} holds where the student was trained and nowhere
else.

\section{Reinforcement learning from the exact reward}
\label{sec:rl}
The exact oracle gives a process reward with no error in it, and the question is whether RL against it
can push the frontier-taught student toward the algorithm-taught one. We warm-start from Frontier-148, since Alg-2000 sits at the reward ceiling and yields no
within-group variance to learn from, and train with REINFORCE and a group-relative baseline over four
rollouts per context on fresh contexts, for 25 steps. Table~\ref{tab:main} gives the held-out result and Appendix~\ref{app:rlfigs} the training-time evaluations.

\paragraph{Reward shape.}
Our first reward weighted sufficiency and recall above exact minimality and precision, with small
terms for answering and a per-probe cost. Within 25 steps it
produced padding. Reward rose by 0.012 while exact minimality fell from 58\% to 33\% and
the surplus grew from 0.5 to 2.0 features (Appendix~\ref{app:rl}). Citing an extra feature raises sufficiency and recall and is charged less than it earns. The
verifier was exact throughout, so the hacking is a property of the reward's shape alone, the
distinction \citet{rubrichacking2026} draw between verifier failure and rubric design in its sharpest form. The corrected reward makes exact
minimality dominant and charges per surplus feature, so padding is strictly worse than the minimal
answer (Appendix~\ref{app:rl} has both in full).

\paragraph{Accuracy does not move and strategy inverts.}
Under the corrected reward, the policy's exact minimality on the 24 fresh contexts used for
training-time evaluation rose from 63\% to 83\%. On the 64 held-out episodes of
Table~\ref{tab:main} it is 56\% against its warm start's 53\%, a difference of three points
($p=0.80$). The training-distribution gain did not transfer, and we report the held-out number as the
result. What did change is how the student spends its budget. Frontier-148 probes 3.5 times and checks
25.5 times per episode. After RL it probes zero times and checks 30 times, the full budget, on every
one of 112 episodes, and commits on its own rather than being cut off. The exact reward turned an
explorer into an exhaustive verifier with the deletion algorithm's query shape, all sufficiency and
no exploration, but without the algorithm's judgment about which check to run next, and so without
its accuracy.

\paragraph{Two warm starts, two outcomes.}
A second run from an independently trained copy of Frontier-148, under the identical reward and
hyperparameters, diverged instead, and exact minimality fell to 6\% on the held-out set with a surplus
of 3.5 features (Appendix~\ref{app:rl}). Its reward fell too, so this is not hacking but a policy
that lost the procedure. One RL run from one warm start is not evidence of a direction, and we have
two runs pointing opposite ways. Both share the strategy inversion, and neither improves the answer.
The third case is the ceiling: RL from Alg-2000 found zero within-group variance on all 21 of its
steps and never took a gradient. Across all three, the exact reward measured what training did, and
in this setting training changed the procedure without improving the answer.

\begin{table*}[t]
  \centering
  \caption{\textbf{Paired tests on exact minimality, GlassGate.} Each row is computed on the episodes
  both systems ran; $p$ is a two-sided sign-flip permutation test with $20{,}000$ draws.}
  \label{tab:pairs}
  \small
  \begin{tabular}{lrrr}
\toprule
comparison (a $-$ b) & $\Delta$Exact & $n$ & $p$ \\
\midrule
algorithm $\times$2000 vs frontier $\times$148 teacher & +0.391 & 64 & $<$.001 \\
algorithm vs frontier teacher, both $\times$148 & +0.156 & 64 & 0.11 \\
algorithm $\times$2000 vs $\times$148 & +0.234 & 64 & 0.003 \\
best student vs Haiku 4.5, same instruction & +0.109 & 64 & 0.12 \\
best student vs Sonnet 5, same instruction & +0.359 & 64 & $<$.001 \\
frontier-taught student vs Haiku 4.5, same instruction & $-$0.281 & 64 & $<$.001 \\
best student vs the deletion ceiling & $-$0.078 & 64 & 0.06 \\
RL from the exact reward vs its warm start & +0.031 & 64 & 0.80 \\
Haiku 4.5 vs Flash-Lite & +0.425 & 120 & $<$.001 \\
Sonnet 5 vs Haiku 4.5 & $-$0.083 & 120 & 0.06 \\
best student vs single-flip heuristic & +0.406 & 64 & $<$.001 \\
frontier-taught student vs single-flip & +0.016 & 64 & 1.00 \\
RL student vs single-flip & +0.047 & 64 & 0.72 \\
\bottomrule
\end{tabular}

\end{table*}

\section{Related work}
\label{sec:related}
\paragraph{Explaining by querying.}
\citet{aiscientists2025} have LLMs recover a hidden program, formal language or equation by
querying it, and find that models plateau under passive observation and improve when allowed to
intervene. \citet{oraclebench2025} benchmark the same protocol across 96 hidden functions, and
\citet{codearc2025} fine-tune an 8B model on traces from a teacher with access to the hidden function,
the closest precedent for training the querier. Our setting shares the premise and differs in the object. Their
explainer hypothesises a whole function; ours explains one decision and is scored exactly against
every minimal sufficient reason of that decision, and we train the querier rather than only
prompting it. \citet{decisionboundaries2025} show self-generated counterfactuals are valid but not minimal, the failure our \texttt{check} tool makes measurable. The protocol is query learning \citep{angluin1988queries,weiss2018extracting}: \texttt{decide} is a membership query, \texttt{check} a sufficiency query, and unlike L$^*$ there is no equivalence oracle.

\paragraph{Formal explanations.}
The minimal sufficient reason is the abductive explanation of \citet{ignatiev2019abduction}, the prime implicant of \citet{shih2018symbolic}, and the sufficient reason of \citet{darwiche2020reasons}, whose term we adopt; \citet{marquessilva2022delivering} argue formal XAI should deliver it. Query-based methods for related objects predate LLMs: Anchors \citep{ribeiro2018anchors} finds
high-precision sufficient conditions of a black box by sampling, \citet{blanc2021provably} give
query-efficient algorithms with guarantees, and the deletion procedure we use as teacher is the
standard extractor for minimal unsatisfiable cores \citep{junker2004quickxplain}. Recent work computes these for black-box vision models by query,
\citet{miss2026} for point clouds and \citet{optimusprime2026} at the concept level. Those are
algorithms, and our question is whether a language model can be taught to run one. An exact oracle replaces the faithfulness proxies of \citet{atanasova2023faithfulness} and \citet{chen2024models}, which \citet{bonafide2026} show do not measure faithfulness.

\paragraph{Verifiers and reward shape.}
Reward hacking under RLVR is documented by \citet{gamingverifiers2026} and
\citet{compositerewards2025}, and \citet{rubrichacking2026} separate verifier failure from rubric
design. \citet{verificationerror2026} and \citet{fuzzingverifiers2026} degrade or fuzz the verifier
to study the first. We hold the
verifier exact, so what remains is the second, and the padding we observe under a sufficiency-heavy
reward is the pattern \citet{cherrl2026} reproduce and detect under rubric rewards, here with verifier
error at zero. \citet{tooluse2026} report RL collapsing tool use while task accuracy holds, which is
the same shape as our strategy inversion; \citet{yue2025boundary} report the same
shape for RLVR more broadly. Process reward models for agents
\citep{agentprm2025} supply dense signal from a learned scorer, and ours is dense and exact, which is
what lets us attribute the strategy inversion to the reward rather than to scorer noise.

\paragraph{Teachers.}
Training small models on traces of a symbolic search is established
\citep{sos2024,guidedsos2024,searchformer2024,procedurecloning2022}, and \citet{beyondsemantics2025} show
students reproduce trace form without its semantics, the mechanism behind our off-distribution result and the
small-model form of the gap \citet{gudibande2023imitating} report.
Distillation from frontier traces is surveyed by \citet{opdsurvey2026}; the closest precedent for a
verifier-guided variant is \citet{aletheia2026}, whose traces carry backtracking where ours carry only a
fixed schedule of checks. We did not find a prior comparison of an
algorithmic against a frontier teacher on identical exact ground truth, which is what
Section~\ref{sec:teaching} runs.

\section{Limitations}
\label{sec:limits}
Alg-148 and Frontier-148 have three training seeds each (Appendix~\ref{app:seeds}); every other arm has one,
so intervals elsewhere are over episodes. Three seeds is also the floor for a seed-level test, since
complete separation of 3 against 3 gives an exact permutation $p$ of 0.10, so the teacher-kind claim
rests on the within-seed paired tests and the non-overlap. One student size and one base model were
trained. GlassGate is a single hand-written policy, and the student loses to Haiku on all three public policies it was
scored on. Sonnet's numbers are bounded by a harness rule, two malformed replies end the episode, that
no other model triggered; a different rule would raise Sonnet and change nothing else. The oracle needs
an enumerable context space, bounding the setting to categorical policies.

\section{Conclusion}
\label{sec:conclusion}
With the verifier exact, every result above is measured against ground truth. The frontier finds a
minimal sufficient reason about half the time with the tools alone and three quarters once
instructed. An algorithm can teach a 4B model to match it where a frontier model cannot, but only on
the gate it was taught on; off it, sufficiency travels and minimality does not. An exact reward
changed the strategy in both RL runs and moved accuracy in opposite directions. In the supervised and
RL results alike, training installed the shape of a procedure without the judgment inside it, and the
exact verifier is what let us tell the two apart.

\section*{Reproducibility}
One script generates every number from per-episode records, restricting each comparison to the
episodes the compared systems share and printing $n$. The policy, the exact oracle, the deletion
algorithm, the training scripts and the per-episode records will be released on publication. Settings
are in Section~\ref{sec:setting} and Appendix~\ref{app:repl}.

\section*{Use of AI assistants}
LLMs are the subject of this work: the frontier models measured, the Opus and Sonnet traces used as
one teacher, and the trained 4B policies. The authors also used LLM assistance for editing text and
for coding, under their direction. Every result comes from the exact oracle and the deterministic
evaluation pipeline, and the authors take full responsibility for the content.

\bibliography{references}

\appendix
\setcounter{topnumber}{3}\setcounter{bottomnumber}{2}\setcounter{totalnumber}{5}
\renewcommand{\topfraction}{0.95}\renewcommand{\bottomfraction}{0.9}
\renewcommand{\textfraction}{0.05}\renewcommand{\floatpagefraction}{0.8}
\renewcommand{\dbltopfraction}{0.95}\renewcommand{\dblfloatpagefraction}{0.8}
\setcounter{dbltopnumber}{3}

\section{Replication details}
\label{app:repl}

\subsection{Hyperparameters and compute}
\label{app:hparams}
Tables~\ref{tab:hparams} and~\ref{tab:compute} give every training and evaluation setting and the compute used.

\begin{table*}[t]
  \centering
  \caption{Training and evaluation settings.}
  \label{tab:hparams}
  \scriptsize
  \begin{tabular}{ll}
\toprule
setting & value \\
\midrule
student base model & Qwen3-4B, chat template, thinking disabled \\
adapter & LoRA rank 16, $\alpha=32$, dropout 0, on q/k/v/o and gate/up/down projections \\
SFT steps, batch, learning rate & 600, 2, $10^{-5}$ (AdamW) \\
SFT precision & fp16 with dynamic loss scaling (T4 has no bf16) \\
teacher A data & 2000 (A) or 148 (A$'$) deletion-algorithm episodes, one \texttt{check} per feature \\
teacher B data & 148 Opus 5 / Sonnet 5 episodes with both tools, full trace \\
evaluation & budget 30, both tools, shrink instruction, greedy decoding, 220 new tokens per turn \\
RL algorithm & REINFORCE, group-relative baseline, 4 rollouts per context \\
RL learning rate, temperature & $5\times10^{-6}$, 0.8 \\
RL budgets & 15 turns during training, 30 at evaluation \\
RL contexts & fresh random contexts each step; 24 fixed held-out contexts for the training-time curve \\
RL steps per leg, total & 25, 25 (one leg reported) \\
RL safeguards & pre-flight: 3 groups, abort if max std $<0.02$; breaker after 10 zero-variance steps \\
frontier models & Gemini 3.5 Flash-Lite, Claude Haiku 4.5, Claude Sonnet 5, Claude Opus 5; thinking off \\
statistics & paired sign-flip permutation, 20{,}000 draws; bootstrap CIs, 2{,}000 resamples \\
\bottomrule
\end{tabular}

\end{table*}
\begin{table*}[t]
  \centering
  \caption{Compute and spend. GPU time is on Kaggle's free T4 tier across two accounts.}
  \label{tab:compute}
  \scriptsize
  \begin{tabular}{lrl}
\toprule
stage & cost & notes \\
\midrule
SFT, three students & 3 $\times$ 44 min T4 & 4.1 s/step \\
student evaluation, 112 episodes & $\approx$ 1.8 h T4 per adapter & 60 s/episode with greedy decoding \\
RL, one 25-step leg & 3.1 h T4 & 310--340 s/step plus a 24-episode evaluation \\
RL adapter evaluation, 112 episodes & 3.4--4.7 h T4 per adapter & the RL policies use the full budget \\
frontier episodes, Gemini & \$3.35 & 10{,}190 calls \\
frontier episodes, Anthropic API & \$3.62 & remainder of Haiku and Sonnet arms \\
frontier episodes, Claude CLI & subscription & most Haiku and Sonnet arms \\
total GPU & $\approx$ 30 T4-hours & two free Kaggle accounts \\
\bottomrule
\end{tabular}

\end{table*}

\subsection{Reward definitions}
\label{app:rl}
Both rewards are functions of the committed set $S$ and the episode's metrics. Reward v1, which
produced padding, was
\begin{align*}
r_1 &= 0.10\,[\text{answered}] + 0.05\,[\text{not forced}]\\
    &\quad + 0.25\,\text{CRec} + 0.15\,\text{CPrec} + 0.25\,[\text{Suff}]\\
    &\quad + 0.20\,[\msr] - 0.01 \cdot \#\text{probes}.
\end{align*}
Reward v3, used for the reported runs, is
\begin{align*}
r_3 &= 0.05\,[\text{answered}] + 0.05\,[\text{not forced}]\\
    &\quad + 0.10\,\text{CRec} + 0.15\,\text{CPrec} + 0.15\,[\text{Suff}]\\
    &\quad + 0.50\,[\msr] - 0.05 \cdot \text{surplus}\\
    &\quad - 0.005 \cdot \#\text{probes}.
\end{align*}
Under $r_3$ the exact minimal answer scores $1.0$ and any padded sufficient answer strictly less.
Each step samples four rollouts per context on fresh contexts; the baseline is the group mean.
A run stops after ten consecutive steps with zero within-group variance.

\subsection{The full policy}
\label{app:policy}
Table~\ref{tab:policy} is the complete GlassGate rule set, generated from the policy source. Rules are
checked in order and the first match decides, so \texttt{R9} fires only when no BLOCK rule matched.

\begin{table*}[t]
  \centering
  \caption{The nine GlassGate rules. \textbf{literals} is the number of feature-value conditions in
  the antecedent, which bounds the size of a block's minimal sufficient reason at three.}
  \label{tab:policy}
  \scriptsize
  \begin{tabular}{@{}llccp{0.52\textwidth}@{}}
\toprule
id & name & decision & literals & condition \\
\midrule
\texttt{R1} & \texttt{credential-egress} & BLOCK & 3 & BLOCK if the payload contains credentials, the target domain is public.web or unknown.xyz, and the tool is an outbound channel (email.send, http.post, or files.write). \\
\texttt{R2} & \texttt{payments-elevation} & BLOCK & 2 & BLOCK if the tool is payments.transfer and the session auth\_level is not elevated. \\
\texttt{R3} & \texttt{destructive-role} & BLOCK & 3 & BLOCK if the declared intent is destructive, the user\_role is not admin, and the tool can modify or delete data (files.delete, files.write, or db.write). \\
\texttt{R4} & \texttt{unauthenticated-write} & BLOCK & 2 & BLOCK if the session auth\_level is none and the requested scope\_level is write or admin. \\
\texttt{R5} & \texttt{pii-external-share} & BLOCK & 3 & BLOCK if the payload contains PII, the declared intent is external\_share, and the target domain is not internal.corp. \\
\texttt{R6} & \texttt{repeat-offender-admin} & BLOCK & 2 & BLOCK if the session has 2 prior policy violations and the requested scope\_level is admin. \\
\texttt{R7} & \texttt{intern-financial-export} & BLOCK & 3 & BLOCK if the user\_role is intern, the declared intent is bulk\_export, and the payload contains financial data. \\
\texttt{R8} & \texttt{unknown-domain-write} & BLOCK & 2 & BLOCK if the target domain is unknown.xyz and the requested scope\_level is write or admin. \\
\texttt{R9} & \texttt{default} & ALLOW & 0 & ALLOW if no earlier rule matched. \\
\bottomrule
\end{tabular}

\end{table*}

\subsection{Prompts}
\label{app:prompts}
The explainer prompt (Figure~\ref{fig:prompts}) is assembled from the blocks shown, in order, followed
by the request context, the decision, the feature domains, the log of actions so far, and the
action-format block for the remaining budget. The shrink instruction is present for every student and
for the shrink arms of Section~\ref{sec:frontier}. The algorithm teacher's traces use the identical
prompt with the deletion algorithm's actions as completions; the frontier teacher's are Opus 5 and
Sonnet 5 transcripts under the identical prompt, unedited.

\begin{figure*}[t]
\input{generated/prompts}
\caption{The prompt blocks given to every explainer.}
\label{fig:prompts}
\end{figure*}

\subsection{A worked episode}
\label{app:example}
Figure~\ref{fig:example} is one GlassGate episode on which Alg-2000 returned the exact minimal reason
and Sonnet 5 did not, with the deletion algorithm's ten checks that Alg-2000 reproduces.

\begin{figure*}[t]
\input{generated/trajectory_example}
\caption{A worked GlassGate episode: the deletion algorithm's trace and Sonnet 5's trace, which ends in a
sufficient but non-minimal answer.}
\label{fig:example}
\end{figure*}

\section{Supplementary tables and figures}
\label{app:figs}
\label{app:offdist}
\label{app:failure}
\label{app:frontier}
\label{app:rlfigs}
\label{app:seeds}
\label{app:ci}
Tables and figures referenced from the main text.

\begin{figure*}[p]
  \centering
  \includegraphics[width=0.66\textwidth]{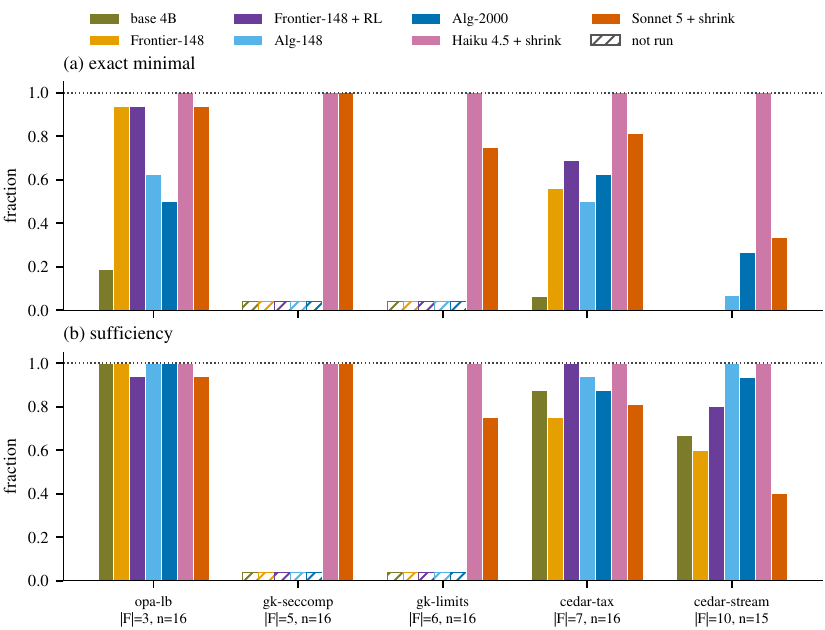}
  \caption{Every system on every public policy, shared episodes per policy. (a) Exact minimality.
  (b) Sufficiency. The dotted line is the deletion algorithm at 1.0. The two Gatekeeper policies were
  run by the frontier models only.}
  \label{fig:perpolicy}
\end{figure*}

\begin{figure*}[p]
  \centering
  \includegraphics[width=0.8\textwidth]{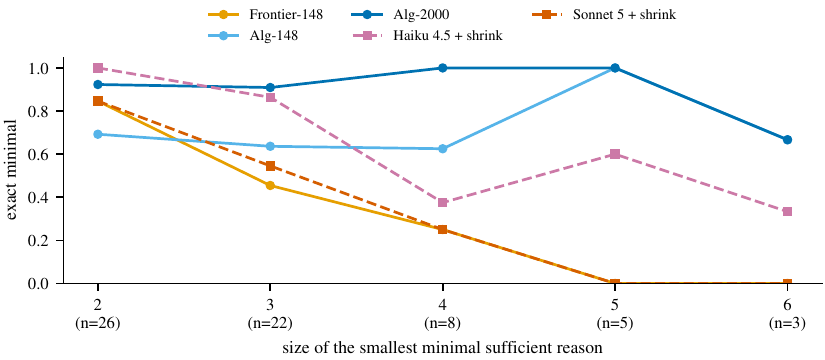}
  \caption{Exact minimality on GlassGate by the size of the smallest MSR, on the 64 episodes all
  systems share. Episode counts per size are in the tick labels. Every episode whose smallest reason
  exceeds three features is an allow. The untrained base is zero at every size and the RL student
  tracks Frontier-148, so both are omitted.}
  \label{fig:bysize}
\end{figure*}

\begin{figure*}[p]
  \centering
  \includegraphics[width=0.85\textwidth]{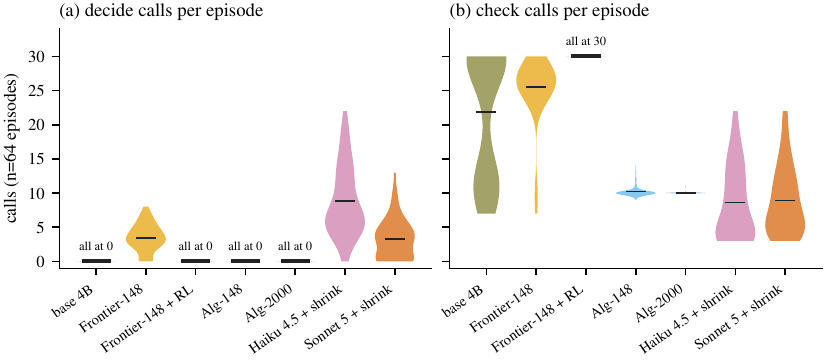}
  \caption{How each system spends its thirty-turn budget on the 64 shared GlassGate episodes.
  Violins are the per-episode distribution; the bar is the mean.}
  \label{fig:budget}
\end{figure*}

\begin{table*}[p]
  \centering
  \caption{All frontier arms with both tools and budget 30, every domain, shared episodes within each
  domain block. Surplus is the mean number of cited features over the smallest MSR. Within each domain
  the first three rows are on the same episodes; Opus is on the subset of those it answered.}
  \label{tab:frontier}
  \scriptsize
  \begin{tabular}{llrccccccc}
\toprule
domain & model & $n$ & Exact & Suff & CPrec & surplus & probes & checks & forced \\
\midrule
glassgate ($|F|$=10) & Flash-Lite & 120 & 0.158 & 0.942 & 0.688 & +1.98 & 1.5 & 10.9 & 0.00 \\
 & Haiku 4.5 & 120 & 0.583 & 0.750 & 0.960 & +0.10 & 12.5 & 4.2 & 0.00 \\
 & Sonnet 5 & 120 & 0.500 & 0.525 & 0.525 & $-$1.55 & 7.0 & 7.4 & 0.46 \\
 & Opus 5 & 21 & 1.000 & 1.000 & 1.000 & +0.10 & 0.3 & 7.3 & 0.00 \\
\addlinespace[2pt]
opa-loadbalancer ($|F|$=3) & Flash-Lite & 16 & 0.438 & 0.938 & 0.844 & +0.44 & 0.6 & 2.2 & 0.00 \\
 & Haiku 4.5 & 16 & 0.812 & 1.000 & 0.969 & +0.19 & 4.7 & 1.8 & 0.00 \\
 & Sonnet 5 & 16 & 1.000 & 1.000 & 1.000 & +0.00 & 3.1 & 1.5 & 0.00 \\
\addlinespace[2pt]
gk-psp-seccomp ($|F|$=5) & Flash-Lite & 16 & 0.688 & 1.000 & 0.875 & +0.31 & 0.3 & 2.2 & 0.00 \\
 & Haiku 4.5 & 16 & 1.000 & 1.000 & 1.000 & +0.00 & 2.6 & 2.0 & 0.00 \\
 & Sonnet 5 & 16 & 1.000 & 1.000 & 1.000 & +0.00 & 1.7 & 1.0 & 0.00 \\
\addlinespace[2pt]
gk-container-limits ($|F|$=6) & Flash-Lite & 16 & 0.312 & 1.000 & 0.796 & +1.12 & 0.7 & 3.9 & 0.00 \\
 & Haiku 4.5 & 16 & 0.688 & 1.000 & 0.937 & +0.44 & 4.2 & 1.5 & 0.00 \\
 & Sonnet 5 & 16 & 0.875 & 1.000 & 0.979 & +0.25 & 2.8 & 1.6 & 0.00 \\
\addlinespace[2pt]
cedar-tax-preparer ($|F|$=7) & Flash-Lite & 16 & 0.500 & 1.000 & 0.823 & +0.75 & 0.4 & 2.4 & 0.00 \\
 & Haiku 4.5 & 16 & 0.562 & 1.000 & 0.984 & +0.50 & 5.2 & 2.1 & 0.00 \\
 & Sonnet 5 & 16 & 0.875 & 0.875 & 0.875 & $-$0.31 & 3.2 & 3.6 & 0.12 \\
\addlinespace[2pt]
cedar-streaming-service ($|F|$=10) & Flash-Lite & 16 & 0.000 & 0.688 & 0.751 & +1.94 & 3.5 & 11.8 & 0.00 \\
 & Haiku 4.5 & 16 & 0.438 & 0.750 & 0.844 & +0.25 & 11.2 & 9.6 & 0.06 \\
 & Sonnet 5 & 16 & 0.375 & 0.500 & 0.545 & $-$0.62 & 7.6 & 9.6 & 0.38 \\
\addlinespace[2pt]
\bottomrule
\end{tabular}

\end{table*}

\begin{figure*}[p]
  \centering
  \includegraphics[width=\textwidth]{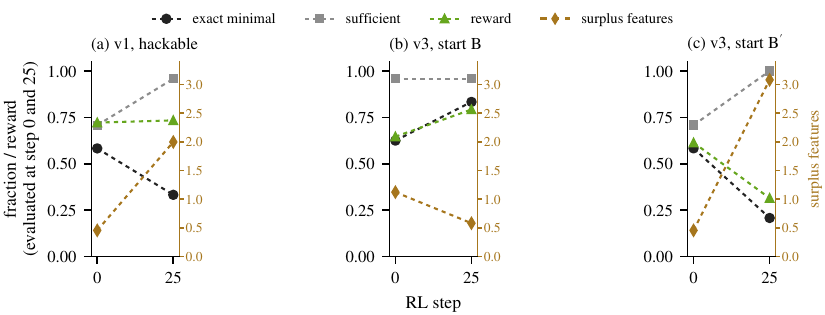}
  \caption{RL from the exact reward, evaluated at step 0 and step 25 on 24 fixed training-time
  contexts; lines connect the two evaluations. (a) Reward $r_1$ is hacked by padding: reward rises
  while exact minimality falls and surplus goes from 0.5 to 2.0. (b) Reward $r_3$ from Frontier-148
  raises exact minimality on these contexts, a gain that does not survive on the 64 held-out episodes
  of Table~\ref{tab:main}. (c) The same reward from a second copy of Frontier-148 diverges. Surplus
  uses the brown right axis.}
  \label{fig:rl}
\end{figure*}

\begin{table*}[p]
  \centering
  \caption{Three training seeds each for the matched-size arms, Alg-148 (A$'$) and Frontier-148 (B), on the 64 GlassGate episodes and the 48
  public-policy episodes the students ran. Seed 0 is the run reported in Table~\ref{tab:main}.
  Within-seed paired differences on GlassGate are $+16$ ($p=0.11$), $+47$ and $+61$ points (both $p<10^{-4}$).}
  \label{tab:seeds}
  \small
  \begin{tabular}{llcccc}
\toprule
teacher & seed & \multicolumn{2}{c}{GlassGate} & \multicolumn{2}{c}{real policies} \\
\cmidrule(lr){3-4}\cmidrule(lr){5-6}
 & & Exact & Suff & Exact & Suff \\
\midrule
algorithm $\times$148 (A$'$) & 0 & 0.688 & 1.000 & 0.396 & 0.979 \\
 & 1 & 0.875 & 0.938 & 0.562 & 0.958 \\
 & 2 & 0.969 & 0.969 & 0.438 & 0.958 \\
 & mean & \textbf{0.844} (sd 0.143) & & \textbf{0.465} & \\
\addlinespace[2pt]
frontier $\times$148 (B) & 0 & 0.531 & 0.781 & 0.500 & 0.771 \\
 & 1 & 0.406 & 0.734 & 0.583 & 0.833 \\
 & 2 & 0.359 & 0.656 & 0.542 & 0.792 \\
 & mean & \textbf{0.432} (sd 0.089) & & \textbf{0.542} & \\
\addlinespace[2pt]
\bottomrule
\end{tabular}

\end{table*}

\begin{table*}[p]
  \centering
  \caption{Per-system means with 95\% bootstrap intervals over episodes (2{,}000 resamples), on each
  system's full episode set. Real-policy rows for the students pool the three policies they ran.}
  \label{tab:ci}
  \scriptsize
  \begin{tabular}{lrcccc}
\toprule
system & $n$ & Exact [95\% CI] & Suff [95\% CI] & CPrec & surplus \\
\midrule
Alg-2000, GlassGate & 64 & 0.922 [0.84, 0.98] & 0.984 [0.95, 1.00] & 0.973 & +0.38 \\
Alg-148, GlassGate & 64 & 0.688 [0.58, 0.80] & 1.000 [1.00, 1.00] & 0.914 & +0.88 \\
Frontier-148, GlassGate & 64 & 0.531 [0.41, 0.66] & 0.781 [0.67, 0.88] & 0.841 & +1.17 \\
Frontier-148 + RL, GlassGate & 64 & 0.562 [0.44, 0.69] & 0.734 [0.62, 0.84] & 0.877 & +0.75 \\
base, GlassGate & 64 & 0.000 [0.00, 0.00] & 0.250 [0.16, 0.36] & 0.430 & +2.42 \\
Haiku 4.5, GlassGate & 120 & 0.583 [0.49, 0.67] & 0.750 [0.67, 0.82] & 0.960 & +0.10 \\
Sonnet 5, GlassGate & 120 & 0.500 [0.41, 0.59] & 0.525 [0.43, 0.61] & 0.525 & $-$1.55 \\
Flash-Lite, GlassGate & 120 & 0.158 [0.10, 0.23] & 0.942 [0.90, 0.98] & 0.688 & +1.98 \\
Opus 5, GlassGate & 21 & 1.000 [1.00, 1.00] & 1.000 [1.00, 1.00] & 1.000 & +0.10 \\
\midrule
Alg-2000, real policies & 48 & 0.479 [0.33, 0.62] & 0.938 [0.85, 1.00] & 0.780 & +1.38 \\
Alg-148, real policies & 48 & 0.396 [0.27, 0.54] & 0.979 [0.94, 1.00] & 0.791 & +1.35 \\
Frontier-148, real policies & 48 & 0.500 [0.35, 0.65] & 0.771 [0.65, 0.88] & 0.785 & +1.02 \\
Frontier-148 + RL, real policies & 48 & 0.542 [0.40, 0.69] & 0.896 [0.79, 0.98] & 0.827 & +1.25 \\
base, real policies & 48 & 0.083 [0.02, 0.17] & 0.854 [0.75, 0.94] & 0.540 & +2.94 \\
\bottomrule
\end{tabular}

\end{table*}

\end{document}